\documentclass{article}

     \usepackage[preprint]{neurips_2022}

\usepackage[utf8]{inputenc} 
\usepackage[T1]{fontenc}    
\usepackage{hyperref}       
\usepackage{url}            
\usepackage{booktabs}       
\usepackage{amsfonts}       
\usepackage{nicefrac}       
\usepackage{microtype}      
\usepackage{xcolor}         

\usepackage{graphicx}
\usepackage[cmex10]{amsmath}
\usepackage{subfig}
\usepackage{array}
\usepackage{booktabs}
\usepackage{multirow}
\usepackage{threeparttable}
\usepackage{amsfonts}
\usepackage{tabularx}
\usepackage{algorithm}
\usepackage{algorithmic}
\usepackage{amsmath}
\usepackage{amssymb}
\DeclareMathOperator*{\argmax}{arg\,max}
\DeclareMathOperator*{\argmin}{arg\,min}
\usepackage{mathtools}
\usepackage{cuted}

\DeclarePairedDelimiter\abs{\lvert}{\rvert}%
\newcolumntype{Y}{>{\centering\arraybackslash}X}
\newcolumntype{s}{>{\hsize=.33\hsize\centering\arraybackslash}X}

\title{Doubly Reparameterized Importance Weighted Structure Learning for Scene Graph Generation}

\author{%
  Daqi~Liu\\
  CVSSP\\
  University of Surrey\\
  Guildford, Surrey GU2 7XH \\
  \texttt{daqi.liu@surrey.ac.uk} \\
   \And
  Miroslaw~Bober\\
  CVSSP\\
  University of Surrey\\
  Guildford, Surrey GU2 7XH \\
  \texttt{m.bober@surrey.ac.uk} \\
  \AND
  Josef~Kittler\\
  CVSSP\\
  University of Surrey\\
  Guildford, Surrey GU2 7XH \\
  \texttt{j.kittler@surrey.ac.uk} \\
}

\begin{document}

\maketitle

\begin{abstract}
 As a structured prediction task, scene graph generation, given an input image, aims to explicitly model objects and their relationships by constructing a visually-grounded scene graph. In the current literature, such task is universally solved via a message passing neural network based mean field variational Bayesian methodology. The classical loose evidence lower bound is generally chosen as the variational inference objective, which could induce oversimplified variational approximation and thus underestimate the underlying complex posterior. In this paper, we propose a novel doubly reparameterized importance weighted structure learning method, which employs a tighter importance weighted lower bound as the variational inference objective. It is computed from multiple samples drawn from a reparameterizable Gumbel-Softmax sampler and the resulting constrained variational inference task is solved by a generic entropic mirror descent algorithm. The resulting doubly reparameterized gradient estimator reduces the variance of the corresponding derivatives with a beneficial impact on learning. The proposed method achieves the state-of-the-art performance on various popular scene graph generation benchmarks.
\end{abstract}

\section{Introduction}

Scene graph generation (SGG) is a structured prediction task aiming to explicitly model objects and their relationships in an image by constructing a corresponding visually-grounded scene graph. Its uses can be found in computer vision tasks such as image captioning [1-3] and visual question answering [4-6]. Currently, the variational Bayesian (VB) [7,8] methodology is generally employed to solve the SGG tasks, in which the variational inference step aims to infer the optimum interpretations $z^*$ from the input images $x$ based on the max aposteriori (MAP) estimation principle, i.e. $z^*=\argmax_{z}p(z|x)$, while the classical cross entropy loss is usually applied to fit the underlying posterior with the ground-truth training samples. Due to the exponential dependencies among the output variables, a computationally tractable variational distribution $q(z)$ is generally used to approximate the underlying computationally intractable posterior $p(z|x)$. For tractability, $q(z)$ in SGG models [9-16] is often assumed to be fully decomposed, and the resulting VB framework is also known as the mean field variational Bayesian (MFVB) [7,8]. The associated inference procedure is also known as the mean field variational inference (MFVI) [7,8].

To leverage the superior feature representation learning capability of modern deep neural networks, the above MFVI step is often formulated using message passing neural network (MPNN) models [15-19], in which two fundamental modules are required: i) visual perception and ii) visual context reasoning [20]. The former aims to locate and instantiate objects and predicates within the input images, while the latter tries to infer their consistent interpretation. In the above formulation, due to the nature of the message passing optimization method, a classical evidence lower bound (ELBO) is often implicitly employed as the variational inference objective. However, the variational approximation inferred from such loose ELBO objective generally underestimates the underlying complex posterior [21], which often leads to inferior generation performance.

To address the above issue, in this paper, we propose a novel doubly reparameterized importance weighted structure learning (DR-IWSL) method, in which a tighter importance weighted lower bound [21] is employed to replace the ELBO as the variational inference objective. A reparameterizable Gumbel-Softmax sampler [22] is applied to draw  $i.i.d.$ samples from the associated distribution to compute the above lower bound. To reduce the gradient variance, we adopt a doubly reparameterized gradient estimator [23] in this paper. The resulting constrained variational inference task is solved by a generic entropic mirror descent algorithm. The proposed DR-IWSL method achieves the state-of-the-art performance on two popular SGG benchmarks: Visual Genome and Open Images V6.

\section{Related Works}

There are two main research directions that are currently investigated in the current SGG literature: 1) designing a  feature extracting structure based on  novel MPNN models [3,11,17,25,26], or different mechanisms of embedding the contextual information into the current MPNN models [12,13,15,19,25]; 2) implementing an unbiased relationship prediction via the following debiasing techniques: instance-level resampling [30], dataset resampling [27-29], bi-level data resampling [16], loss reweighting based on instance frequency [35,36] and knowledge transfer learning [32-34]. Besides the above traditional debiasing methodologies, another approach involves the use of a causal inference model [37] to remove the harmful bias from the good context bias based on the counterfactual causality.

Most of the above SGG models [10,13,16,19,26,38] follow a unified MPNN-based MFVB formulation, in which the classical ELBO is implicitly employed as the variational inference objective. However, the resulting variational approximation derived from the ELBO objective generally underestimates the underlying complex posterior. Unlike the previous SGG models, the proposed DR-IWSL method applies a tighter importance weighted lower bound [21] as the variational inference objective, which is computed from multiple samples drawn from a reparameterizable Gumbel-Softmax sampler [22]. Moreover, we employ a doubly reparameterized gradient estimator [23] to reduce the variance of the associated derivatives. Instead of relying on the traditional message passing technique, a generic entropic mirror descent algorithm is used to solve the resulting constrained variational inference task.

\section{Proposed Methodology}

To convey effectively the innovative features of the proposed methodology, in this section, we first formulate the problem, and define the applied scoring function. The presentation then proceeds by motivating the employed Gumbel-Softmax sampler and describing the proposed doubly reparameterized importance weighted structure learning method. The adopted entropic mirror descent method is discussed in the last subsection.

\subsection{Problem Formulation}

In the current SGG approaches, the output scene graph consists of a list of intertwined semantic triplet structures, each of which constrains three key graph building components:  object, subject,  and predicate. In particular, the relationship between two interacting instances (object and subject) is referred as a predicate. Due to the exponential dependencies among the structured output variables in SGG tasks, a direct computation of the underlying posterior is generally computationally intractable. For this reason, the classical variational inference (VI) technique is often applied to approximate the above posterior. For tractability, the mean field variational inference (MFVI) [7,8] is commonly used in such SGG tasks, in which the variational distribution is often assumed to be fully decomposable. Equipped with the classical cross entropy loss in the associated variational learning step, current SGG models can be formulated in a corresponding mean field variational Bayesian (MFVB) framework [7,8]. The above MFVI step is predominantly modelled by a message passing neural network (MPNN) [15-19], consisting of two fundamental modules: visual perception and visual context reasoning. In fact, a MPNN-based MFVB framework has became the de facto state-of-the-art method for SGG.  

Specifically, given an input image $x$, the visual perception module aims to generate a set of instance region proposals $b_i^o\in \mathbb{R}^4,i=1,...,m$, and a set of predicate region proposals $b_j^p\in \mathbb{R}^4,j=1,...,n$, where $m$ and $n$ represent the number of instances/predicates detected in the input image. By applying a ROI pooling on the feature maps generated from the visual perception module, one can extract the associated fixed-sized latent feature representation sets $y_i^o\in \mathbb{R}^d,i=1,...,m$ and $y_j^p \in \mathbb{R}^d, j=1,...,n$ from the corresponding input image patch sets $x_i^o,i=1,...,m$ and $x_j^p,j=1,...,n$. Given a set of object classes $\mathcal{C}$ and a set of relationship categories $\mathcal{R}$, a visual context reasoning module is required to infer the resulting instance/predicate interpretation sets $z_i^o\in \mathcal{C},i=1,...,m$ and $z_j^p\in \mathcal{R},j=1,...,n$ from the above latent feature representation sets.

Traditionally, ELBO is routinely applied as the variational inference objective in the above MPNN-based MFVB models. However, the oversimplified variational approximation inferred from the ELBO objective generally underestimates the underlying complex posterior [21], which often leads to inferior detection performance. To this end, in this paper, we propose a novel doubly reparameterized importance weighted structure learning method, which employs a tighter importance weighted lower bound [21] as the variational inference objective, and utilizes a doubly reparameterized gradient estimator [23] to approximate the associated derivatives. aiming to reduce the estimator variance. 

\subsection{The Scoring Function}

Generally, SGG tasks can be formulated using probabilistic graphical models, e.g. a conditional random field (CRF) [40]. A non-negative scoring function $s_{\theta}(x,z)$ is often applied to measure the similarity or compatibility between the input variable $x$ and the output variable $z$, where $\theta$ is used to parameterize the scoring function. The associated log scoring function is often computed as follows:
\begin{equation}
 logs_{\theta}(x,z)=-\sum_{r\in R}\psi_r(x_r,z_r)
\end{equation}
where $r$ represents a clique within a clique set $R$ (defined by the associated graph structure), $\psi_r$ is a corresponding potential function. Two types of potential functions are commonly used in current SGG models: the unary potential function $\psi_u$ and the pairwise or binary potential function $\psi_b$.

However, the above formulation often ignores the informative global contextual information. To this end, we compute a latent global feature representation $y^g\in \mathbb{R}^d$ from the global region proposal $b^g$, where $b^g$ is obtained by the union of all the associated instance/predicate region proposals in the input image. Correspondingly, $x^g$ is the relevant global image patch of $b^g$, and $z^g$ is its interpretation. 
With the above definitions, by adding two types of pairwise potential terms $\psi_{b}^p(x_j^p, x^g,z_j^p, z^g)$ and $\psi_{b}^o(x_i^o, x^g,z_i^o, z^g)$, one can incorporate the global contextual information into the following applied log scoring function:
\begin{equation}
\begin{split}
logs_{\theta}(x,z)= -\displaystyle\sum_{j=1}^{n}[\psi_u^p(x_j^p, z_j^p)+
\displaystyle\sum_{i\in N(j)}\psi_{b}^p(x_i^o, x_j^p,z_i^o, z_j^p)
+\psi_{b}^p(x_j^p, x^g,z_j^p, z^g)] -\\
\displaystyle\sum_{i=1}^{m}[\psi_u^o(x_i^o, z_i^o)+
\displaystyle\sum_{j\in N(i)}\psi_{b}^o(x_i^o, x_j^p,z_i^o, z_j^p) +\displaystyle\sum_{l\in N(i)}\psi_{b}^o(x_i^o, x_l^o,z_i^o, z_l^o)+
\psi_{b}^o(x_i^o, x^g,z_i^o, z^g)]
\end{split}
\end{equation}
where the superscripts $o$, $p$, $g$ represent the object, the predicate and the global context, respectively. $N(i)$ is the set of neighbouring nodes around the target $i$, and the latent feature representations $y$ are implicitly embedded in the above formulation.

\subsection{Gumbel-Softmax Sampler}

Due to the inability to backpropagate the corresponding gradients, the output discrete variables are rarely applied in stochastic neural networks. To this end, rather than producing non-differentiable samples from a categorical distribution, we employ a reparameterizable Gumbel-Softmax sampler [22] to generate differentiable samples.

Suppose $z$ is the interpretation of a potential region proposal defined in terms of  a categorical variable with the class probabilities $\pi^1,..., \pi^v$ (where $v$ is the vocabulary size). It is essentially encoded as $v$-dimensional one-hot vector taking values on the corner of the $(v-1)$-dimensional simplex, $\Delta^{v-1}$. Given a $v$-dimensional Gumbel noise $\sigma$, the corresponding $i$-th element of the output variable $z^i$ is computed as follows:
\begin{equation}
 z^i=g_{\pi}(\sigma^i)=\frac{exp((log(\pi^i)+\sigma^i)/\tau)}{\sum_{j=1}^{v}exp((log(\pi^j)+\sigma^j)/\tau)},\; for\; i=1,...,v
\end{equation}
where $g_{\pi}$ represents the reparameterization function and $\tau$ is the softmax temperature. With the Gumbel-Softmax sampler, the output samples become one-hot vectors when annealing $\tau$ to zero. To avoid exploding gradients, $\tau$ is often annealed to a relatively low temperature instead of zero.

\subsection{Doubly Reparameterized Importance Weighted Structure Learning}

To avoid underestimating the underlying complex posterior, in this paper, we employ a tighter lower bound $\mathcal{L}_s$ based on $s$-sample importance weighting [21] to replace the classical ELBO as the variational inference objective. Such importance weighted lower bound $\mathcal{L}_s$ is essentially an unbiased estimator of the log partition function $logs_{\theta}(x)$ (when $s$ reaches infinity), which is defined as follows:
\begin{equation}
  \mathcal{L}_{s}=\mathbb{E}_{z_1,...,z_s \sim q(z)}[log\frac{1}{s}\sum_{i=1}^{s}\frac{s_{\theta}(x,z_i)}{q(z_i)}] \leq logs_{\theta}(x)
\end{equation}
where $s$ represents the number of samples, in which each $z_i$ is an $i.i.d.$ random sample drawn from $q(z)$. $w_i=\frac{s_{\theta}(x,z_i)}{q(z_i)}$ is also known as the importance weight. $\mathcal{L}_{s}$ is at least as tight as the ELBO, and its tightness improves with the number of samples [21].

For tractability, the applied variational distribution $q(z)$ is generally assumed to be fully decomposed as:
\begin{equation}
\begin{split}
q(z)=\displaystyle\prod_{i=1}^{m}q^o_i(z_i^o)\displaystyle\prod_{j=1}^{n}q^p_j(z_j^p)
\end{split}
\end{equation}
where $q^o_i(z_i^o) \in \Delta ^{v_o-1}$ and  $q^p_j(z_j^p) \in \Delta ^{v_p-1}$ are local variational approximations of the objects and predicates in the output scene graph, respectively. $v_o$ and $v_p$ are the sizes of vocabularies for the objects and predicates, respectively. In such MFVI scenario, the original MAP inference can be transferred into a corresponding marginal inference task, which may not be the case in general. 

Given a potential region proposal $b_i$, its corresponding local log marginal posterior $logp_{\theta}(z_i|x_i)=logs_{\theta}(x_i,z_i)-logs_{\theta}(x_i)$ requires us to compute the local log marginal scoring function $logs_{\theta}(x_i,z_i)=\sum_{z\backslash z_i}logs_{\theta}(x_i,z)$ and the computationally intractable log partition function $logs_{\theta}(x_i)$. In this paper, variable elimination techniques are applied to approximate $logs_{\theta}(x_i,z_i)$. Specifically, for a potential instance region proposal $b_i^o$, it is computed as follows:
\begin{equation}
\label{con1}
\begin{split}
logs_{\theta}(x_i^o,z_i^o)\propto -[\psi_u^o(x_i^o, z_i^o)+\sum_{j\in N(i)} m^{op}_{j\to i}+\sum_{l\in N(i)} m^{oo}_{l\to i}+m^{og}_{g\to i}] \\
\psi_u^o(x_i, z_i^o) = h^o_{\theta}(x_i)\cdot z_i^o,\;
m^{op}_{j\to i}=\sum_{z_j^p\in \mathcal{R}}\psi^o_b(x_i^o, x_j^p, z_i^o, z_j^p)=g^{op}_{\theta}(x_i^o, x_j^p)\cdot z_i^o\\
m^{oo}_{l\to i}=\sum_{z_l^o\in \mathcal{C}}\psi^o_b(x_i^o, x_l^o, z_i^o, z_l^o)=g^{oo}_{\theta}(x_i^o, x_l^o)\cdot z_i^o\\
m^{og}_{g\to i}=\sum_{z^g\in \mathcal{G}}\psi^o_b(x_i^o, x^g, z_i^o, z^g)=g^{og}_{\theta}(x_i^o, x^g)\cdot z_i^o\\
\end{split}
\end{equation}
while for a potential predicate region proposal $b_j^p$, it is computed as follows:
\begin{equation}
\label{con2}
\begin{split}
logs_{\theta}(x_j^p,z_j^p)\propto -[\psi_u^p(x_j^p, z_j^p)+\sum_{i\in N(j)} m^{po}_{i\to j}+m^{pg}_{g\to j}]\\
\psi_u^p(x_j, z_j^p) = h^p_{\theta}(x_j)\cdot z_j^p,\;
m^{po}_{i\to j}=\sum_{z_i^o\in \mathcal{C}}\psi^p_b(x_i^o, x_j^p, z_i^o, z_j^p)=g^{po}_{\theta}(x_i^o, x_j^p)\cdot z_j^p\\
m^{pg}_{g\to j}=\sum_{z^g\in \mathcal{G}}\psi^p_b(x^g, x_j^p, z^g, z_j^p)=g^{pg}_{\theta}(x_j^p, x^g)\cdot z_j^p
\end{split}
\end{equation}
where $\cdot$ means an inner product, $z_i^o$ and $z_j^p$ are the output variables for an instance and a predicate, which are generated by a Gumbel-Softmax sampler. In Equations (6) and (7), $\mathcal{G}$ is a relevant global region proposal interpretation set. By combing the visual perception module outputs using  multi-layer perceptrons (MLPs), one can construct the relevant feature representation learning functions $h^o_{\theta}$, $h^p_{\theta}$, $g^{op}_{\theta}$, $g^{oo}_{\theta}$, $g^{og}_{\theta}$, $g^{po}_{\theta}$, $g^{pg}_{\theta}$, which are parameterized by $\theta$. Essentially, these functions would first map the input image patches $x$ into the corresponding feature representations $y\in \mathbb{R}^d$ using the visual perception module, and then obtain the resulting $\mathbb{R}^v$ dimensional feature vector by feeding the relevant $y$ into the corresponding MLP. Most importantly, the MLPs implicitly perform the potential function marginalizations prescribed in Equations (6) and (7). The resulting log score is essentially the inner product of the above $\mathbb{R}^v$ dimensional feature vector and the corresponding $v$-dimensional vector $z$.

More specifically, to approximate the computationally intractable $logs_{\theta}(x_i)$, $s$-sample importance weighted lower bound $\mathcal{L}_s^i$ is employed in this paper to construct the following constrained variational inference task: 
\begin{equation}
\begin{split}
  logs_{\theta}(x_i)\triangleq \max_{\pi_i} \mathcal{L}_s^i&=\max_{\pi_i}\mathbb{E}_{z_{i1},...,z_{is}\sim q_{\pi_i}(z_i)}[log\frac{1}{s}\sum_{j=1}^s\frac{s_{\theta}(x_i,z_{ij})}{q_{\pi_i}(z_{ij})}]\\
  &=\max_{\pi_i}\mathbb{E}_{\sigma_{i1},...,\sigma_{is}\sim u(\sigma_i)}[log\frac{1}{s}\sum_{j=1}^s\frac{s_{\theta}(x_i,z_{ij})}{q_{\pi_i}(z_{ij})}]|_{z_{ij}=g_{\pi_i}(\sigma_{ij})}\\
  &\triangleq \max_{\pi_i}[log\frac{1}{s}\sum_{j=1}^s\frac{s_{\theta}(x_i,z_{ij})}{q_{\pi_i}(z_{ij})}]|_{z_{ij}=g_{\pi_i}(\sigma_{ij}),\; \sigma_{ij \sim u(\sigma_i)}} \;\;\;s.t. \;\;\; \pi_i \in \Delta^{v-1}
\end{split}
\end{equation}
where the local variational approximation $q_i(z_i)$ is set to a Gumbel-Softmax distribution with a categorical probability $\pi_i\in \Delta^{v-1}$. and $z_{i1},...,z_{is}$ represent the $s$ $i.i.d$ samples drawn from $q_{\pi_i}(z_i)$. $\sigma_{ij}$ is $v$-dimensional Gumbel noise drawn from the Gumbel distribution $u(\sigma_i)$, which is fed into the Gumbel-Softmax reparameterization function $g_{\pi_i}$ to explicitly compute the corresponding output sample $z_{ij}$. A Monte Carlo estimator is applied to approximate the expectation $\mathcal{L}_s^i$. Accordingly, the above log probability $logq_{\pi_i}(z_{ij})$ is approximated as follows:
\begin{equation}
 logq_{\pi_i}(z_{ij})\triangleq \lVert \pi_i\cdot z_{ij} \rVert_1 -max(\pi_i)-log \lVert e^{\pi_i-max(\pi_i)}\rVert_1
\end{equation}
where $\lVert. \rVert_1$ represents the $\mathbb{L}_1$ norm while $max(\pi_i)$ is the maximum value of $\pi_i$. 

Generally, naively computing the derivatives $\triangledown_{\pi_i} \mathcal{L}_s^i$ would generate a major problem, as the relevant gradient estimator for the above importance weighted lower bound performs poorly as the number of samples increases [23]. To this end, in this paper, we employ a doubly raparameterized gradient estimator [23] to reduce the variance of the associated derivatives. The estimator is expressed as follows:
\begin{equation}
\triangledown_{\pi_i} \mathcal{L}_s^i \triangleq [\sum_{j=1}^s (\frac{w_{ij}}{\sum_{l=1}^s w_{il}})^2 \frac{\partial {logw_{ij}}}{\partial{z_{ij}}} \frac{\partial {z_{ij}}}{\partial{\pi_i}}]|_{z_{ij}=g_{\pi_i}(\sigma_{ij}),\; \sigma_{ij \sim u(\sigma_i)}}
\end{equation}
where $w_{ij}=\frac{s_{\theta}(x_i,z_{ij})}{q(z_{ij})}$ represents the associated importance weight of the $j$-th sample in the $i$-th region proposal. Such doubly reparameterized gradient estimator has the property that when $q_{\pi_i}(z_i)$ is optimal (exactly the same as the underlying posterior), the estimator vanishes and has zero variance. This property does not hold for the naive gradient estimator [23].

Furthermore, a surrogate logit $\phi$ is constructed to compute the target log marginal posterior $logp_{\theta}(z_i|x_i)$:
\begin{equation}
 \begin{split}
     logp_{\theta}(z_i|x_i)\triangleq \phi+C,\;\;\;
     \phi=logs_{\theta}(x_i,z_i)-\max_{\pi_i}\mathcal{L}_s^i
 \end{split}
\end{equation}
where $C$ is a relevant constant w.r.t. $x_i$ and $z_i$. One can compute $logp_{\theta}(z_i|x_i)$ by ignoring the above constant $C$ based on the $LogSumExp$ trick:
\begin{equation}
\begin{split}
logp_{\theta}(z_i|x_i)\triangleq \phi - log{\lVert e^{\phi} \rVert_1}
\end{split}
\end{equation}
where the optimum interpretation $z_i^*$ of the input region proposal $b_i$ is computed as $z_i^*=\argmax_{z_i}logp_{\theta}(z_i|x_i)$. 
\begin{algorithm}[ht]
 \caption{Doubly Raparameterized Importance Weighted Structure Learning}\label{euclid1}
 \textbf{Input} region proposal $b$, categorical probability $\pi$, number of samples $s$, Gumbel noise distribution $u(\sigma)$, Gumbel-Softmax reparameterization function $g_{\pi}$, learning rate $\alpha$, softmax temperature $\tau$, minimum temperature $\tau_{min}$, temperature annealing rate $\beta$, number of iterations $T$ \\
 \textbf{Output} $\theta$, $\tau$
 \begin{algorithmic}[1]
 \STATE randomly initialize $\theta$ 
 \FOR{iteration $t=1$ to $T$}
 \STATE randomly initialize $\pi$ for $b$
 \STATE draw $s$ Gumbel noise samples $\sigma_{1},...,\sigma_{s}$ from $u(\sigma)$
 \STATE compute $s$ output samples $z_{1},...,z_{s}$ by feeding $\sigma_{1},...,\sigma_{s}$ into $g_{\pi}$
 \STATE compute log importance weight $log\frac{s_{\theta}(x,z)}{q_{\pi}(z)}$ and approximate $\mathcal{L}_s$ via Monte Carlo estimation
 \STATE employ EMD to solve the resulting constrained variational inference task, in which the derivative $\triangledown_{\pi} \mathcal{L}_s$ is approximated via a doubly reparameterized gradient estimator
 \STATE apply the updated $\pi$ to compute the surrogate logit $\phi$ as well as the resulting $logp_{\theta}(z|x)$ 
 \STATE compute $\mathbb{L}(\theta)$ and update $\theta \leftarrow \theta - \alpha \cdot \bigtriangledown_{\theta} \mathbb{L}(\theta)$
 \STATE update $\tau \leftarrow \max(\tau \cdot e^{-\beta \cdot t}, \tau_{min})$
 \ENDFOR
 \end{algorithmic}
 \end{algorithm}

Finally, in the following variational learning step, we employ the classical cross-entropy loss to fit the above $p_{\theta}(z|x)$ with the ground-truth training samples:
\begin{equation}
    \theta^*=\argmin_{\theta}\mathbb{L}(\theta)=\argmin_{\theta}-[\frac{1}{c}\sum_{k=1}^{c}logp_{\theta}(\hat{z_k}|\hat{x_k})]
\end{equation}
where $\mathbb{L}(\theta)$ represents the variational learning objective, $c$ is the number of training images in a mini-batch, $\hat{z_k}$ is the ground-truth scene graph of the input image $\hat{x_k}$. To better illustrate the proposed DR-IWSL method, we summarize its learning steps in Algorithm 1.
\begin{algorithm}[ht]
 \caption{Entropic Mirror Descent}\label{euclid2}
 \textbf{Input} variational distribution $\pi$, importance weighted lower bound $\mathcal{L}_s$, number of iterations $M$, an initial learning rate $\gamma$, a predefined objective $\mathcal{L}_s^p$, a small positive value $\epsilon$\\
 \textbf{Output} optimum $\pi^*$
 \begin{algorithmic}[1]
 \FOR{iteration $i=1$ to $M$}
 \STATE compute the derivative $\bigtriangledown_{\pi}\mathcal{L}_s$ via a doubly reparameterized gradient estimator
 \STATE set learning rate $\gamma = \frac{\gamma}{\sqrt{i}}$
 \STATE end the loop if {$\abs{\mathcal{L}_s-\mathcal{L}_s^p}<\epsilon$}
 \STATE set $\mathcal{L}_s^p=\mathcal{L}_s$
 \STATE compute $r=\gamma\cdot\bigtriangledown_{\pi}\mathcal{L}_s$
 \STATE compute $r=\pi\cdot e^{r-\max(r)}$
 \STATE set $\pi=\frac{r}{\lVert r \rVert_1}$
 \ENDFOR
 \end{algorithmic}
 \end{algorithm}

\subsection{Entropic Mirror Descent}

Unlike the previous SGG models, the proposed DR-IWSL method requires us to solve a constrained variational inference task, that is to maximize the $s$-sample importance weighted lower bound $\mathcal{L}_s^i$, subject to the constraint that the categorical probability $\pi_i$ resides in a $(v-1)$-simplex, as demonstrated in Equation (8). Since the above constraint is a probability simplex, entropic mirror descent (EMD) [24] is chosen to solve the above constrained variational inference problem. Specifically, the negative entropy is applied as a specific function to construct the required Bregman distance [45]. Compared with the traditional projected gradient descent methods [47], EMD generally converges faster due to the utilization of the geometry of the optimization problem [46]. To intuitively illustrate the applied EMD strategy, we summarize the specific training steps in Algorithm 2.

\section{Experiments}

\subsection{Visual Genome}

\textbf{Benchmark:} As the most popular SGG benchmark, Visual Genome [48] consists of 108,077 images with an average of 38 objects and 22 relationships per image. Following the data split protocol [9], the most frequent 150 object categories and 50 predicate classes are selected in this experiment. Furthermore, we split the Visual Genome into a training set ($70\%$) and a test set ($30\%$). For validation, an evaluation set ($5k$) is randomly selected from the training set. Following [50], according to the number of instances in the training split, the relevant categories are divided into three disjoint sets: $head$ (more than $10k$), $body$ ($0.5k\sim 10k$) and $tail$ (less than $0.5k$).

\noindent \textbf{Evaluation Metrics:} Instead of the common Recall$@K$, the mean Recall$@K$ ($mR@K$) is chosen as the evaluation metric in this experiment, since it focuses on the informative predicate categories (e.g. $painted\;on$) with much less training samples compared to the common ones (e.g. $on$). 
We validate the proposed method on three tasks, namely, Predicate Classification (PredCls), Scene Graph Classification (SGCls) and Scene Graph Detection (SGDet). In particular, given an input image with the ground-truth bounding boxes and object labels, PredCls predicts the predicate labels; SGCls predicts the labels for instances and predicates; SGDet generates the resulting scene graph from the input image.

\noindent \textbf{Implementation Details:} Like [37], we choose RexNeXt-101-FPN [51] and Faster-RCNN [39] as the backbone and the object detector, respectively. Following the previous methods, we adopt a step training strategy. Accordingly, we freeze the visual perception module and only train the relevant visual context reasoning module. A bi-level data resampling strategy [16] is applied in this experiment. Specifically, we set the repeat factor $t=0.07$ and the instance drop rate $\gamma_{d}=0.7$. The batch size $bs$ is set to 12, and the learning rate of the SGD optimizer is $0.008\times bs$. The number of samples $s$ is set to $20$ and $5000$ in the variational inference and learning steps, respectively. The experiment is conducted on 4 GeForce RTX 3090 GPU cards.
\begin{table}[!t]
   \begin{threeparttable}
	\renewcommand{\arraystretch}{1.0}
	\caption{A performance comparison on Visual Genome dataset.}
	\centering
    \begin{tabular}{@{\extracolsep{4pt}}*7c@{}}
	\toprule
	{} & \multicolumn{2}{c}{PredCls} & \multicolumn{2}{c}{SGCls} & \multicolumn{2}{c}{SGDet}\\ \cmidrule{2-3} \cmidrule{4-5} \cmidrule{6-7}
	{Method} & {mR@50} & {mR@100} & {mR@50} & {mR@100} & {mR@50} & {mR@100}\\
	\midrule
    RelDN$^{\dagger}$[38]  & $15.8$ & $17.2$ & $9.3$ & $9.6$ & $6.0$ & $7.3$\\
	Motifs[26]  & $14.6$ & $15.8$ & $8.0$ & $8.5$ & $5.5$ & $6.8$\\
	Motifs*[26]  & $18.5$ & $20.0$ & $11.1$ & $11.8$ & $8.2$ & $9.7$\\
	G-RCNN$^{\dagger}$[13]   & $16.4$ & $17.2$ & $9.0$ & $9.5$ & $5.8$ & $6.6$\\
	MSDN$^{\dagger}$[10]   & $15.9$ & $17.5$ & $9.3$ & $9.7$ & $6.1$ & $7.2$\\
    GPS-Net$^{\dagger}$[19]   & $15.2$ & $16.6$ & $8.5$ & $9.1$ & $6.7$ & $8.6$\\
    GPS-Net$^{\dagger *}$[19]   & $19.2$ & $21.4$ & $11.7$ & $12.5$ & $7.4$ & $9.5$\\
    VCTree-TDE[37] & $25.4$ & $28.7$ & $12.2$ & $14.0$ & $9.3$ & $11.1$\\
    BGNN[16]  & $30.4$ & $32.9$ & $14.3$ & $16.5$ & $10.7$ & $12.6$\\
    \textbf{DR-IWSL} & $\mathbf{30.4}$ & $\mathbf{32.3}$ & $\mathbf{17.4}$ & $\mathbf{19.0}$ & $\mathbf{14.6}$ & $\mathbf{16.7}$\\
	\bottomrule
    \end{tabular}
    \begin{tablenotes}
	\item [\textbullet] Note: All the above methods apply ResNeXt-101-FPN as the backbone. $*$ means the re-sampling strategy [30] is applied in this method, and $\dagger$ depicts the results reproduced with the latest code from the authors. A bold font marks the results with the proposed method.
      \end{tablenotes}
    \end{threeparttable}
\end{table} 

\noindent \textbf{Comparisons with State-of-the-art Methods:} As shown in Table 1, the proposed DR-IWSL method outperforms the previous state-of-the-art SGG models by a large margin in the SGCls abd SGDet tasks, and achieves comparable performance with the latest BGNN algorithm in the PredCls task. Since the EMD algorithm converges faster than the traditional message passing technique, the above performance is achieved using  a much fewer training iterations. Furthermore, the proposed method focuses on detecting informative predicates ($body$ and $tail$) rather than the common ones ($head$). 

In order to improve the above informative predicate detection capability further, we enhance the proposed method by a generic balance adjustment (BA) strategy. The resulting novel algorithm, referred to as  DR-IWSL+BA, is compared with three baseline models as shown in Table 2. The BA strategy aims to overcome two types of imbalance, namely, the semantic space imbalance and the training sample imbalance. It involves two procedures: semantic adjustment and balanced predicate learning. The former induces the predictions made by the DR-IWSL method to be more informative by building a relevant transition matrix, while the latter aims to extend the sampling space for the informative predicates. We observe that the proposed DR-IWSL+BA method outperforms the previous state-of-the-art models by a large margin, especially for the PredCls task.
\begin{table}[!t]
   \begin{threeparttable}
	\renewcommand{\arraystretch}{1.0}
	\caption{A comparison of the impact of the balance strategy measured on Visual Genome.}
	\centering
    \begin{tabular}{@{\extracolsep{4pt}}*7c@{}}
	\toprule
	{} & \multicolumn{2}{c}{PredCls} & \multicolumn{2}{c}{SGCls} & \multicolumn{2}{c}{SGDet}\\ \cmidrule{2-3} \cmidrule{4-5} \cmidrule{6-7}
	{Method} & {mR@50} & {mR@100} & {mR@50} & {mR@100} & {mR@50} & {mR@100}\\
	\midrule
	Motifs+BA[34]  & $29.7$ & $31.7$ & $16.5$ & $17.5$ & $13.5$ & $15.6$\\
	VCTree+BA[34]  & $30.6$ & $32.6$ & $20.1$ & $21.2$ & $13.5$ & $15.7$\\
    Transformer+BA[34]  & $31.9$ & $34.2$ & $18.5$ & $19.4$ & $14.8$ & $17.1$\\
    \textbf{DR-IWSL+BA} & $\mathbf{37.7}$ & $\mathbf{40.0}$ & $\mathbf{21.5}$ & $\mathbf{22.7}$ & $\mathbf{16.5}$ & $\mathbf{18.7}$\\
	\bottomrule
    \end{tabular}
    \begin{tablenotes}
	\item [\textbullet] Note: All the above methods apply the same balance adjustment strategy as in [34]. Using bold to identify the proposed method.
      \end{tablenotes}
    \end{threeparttable}
\end{table}

\subsection{Open Images V6}

\textbf{Benchmark:} Open Images V6 [49] is another popular SGG benchmark, which has a superior annotation quality and includes 126,368 training images, 5322 test images and 1813 validation images. The same data processing protocols in [19,38,49] are selected in this experiment.

\noindent \textbf{Evaluation Metrics:} Similar to the evaluation protocols in [19,38,49], we choose the following evaluation metrics in this experiment: the mean Recall$@50$ ($mR@50$), the regular Recall$@50$ ($R@50$), the weighted mean AP of relationships ($wmAP_{rel}$) and the weighted mean AP of phrase ($wmAP_{phr}$). Specifically, as in [19,38,49], the weight metric score is defined as: $score_{wtd}=0.2\times R@50 + 0.4\times wmAP_{rel} + 0.4\times wmAP_{phr}$.

\noindent \textbf{Implementation Details:} Like the experiment in Visual Genome, we choose the same backbone and object detector in this experiment. Moreover, we employ the same step training strategy and bi-level data resampling technique. The batch size $bs$ is set to 12 and an Adam optimizer with the learning rate of $0.0001$ is utilized. The number of samples $s$ is set to $20$ and $5000$ in the variational inference and learning steps, respectively. 

\noindent \textbf{Comparisons with State-of-the-art Methods:} To verify the merits of the proposed DR-IWSL method further, we compare it with various state-of-the-art SGG models in Table 3. For a fair comparison, whenever possible, we reproduce the results of the methods with the author's latest code and, over and above that, we investigate the impact of using  the additional re-sampling strategy proposed in [30] on two of those methods. As shown in Table 3, the proposed DR-IWSL method achieves the state-of-the-art performance on all evaluation metrics for the Open Images V6 dataset.
\begin{table}[!t]
   \begin{threeparttable}
	\renewcommand{\arraystretch}{1.0}
	\caption{A performance comparison on the Open Images V6 dataset.}
	\centering
    \begin{tabular}{@{\extracolsep{4pt}}*6c@{}}
	\toprule
	{Method} & {mR@50} & {R@50} & {wmAP\_rel} & {wmAP\_phr} & {score\_wtd} \\
	\midrule
	RelDN$^{ \dagger}$[38]  & $33.98$ & $73.08$ & $32.16$ & $33.39$ & $40.84$\\
	RelDN$^{\dagger*}$[38]  & $37.20$ & $75.34$ & $33.21$ & $34.31$ & $41.97$ \\
    VCTree$^{\dagger}$[15] & $33.91$ & $74.08$ & $34.16$ & $33.11$ & $40.21$ \\
    G-RCNN$^{\dagger}$[13]   & $34.04$ & $74.51$ & $33.15$ & $34.21$ & $41.84$ \\
	Motifs$^{\dagger}$[26]  & $32.68$ & $71.63$ & $29.91$ & $31.59$ & $38.93$ \\
    VCTree-TDE$^{\dagger}$[37]  & $35.47$ & $69.30$ & $30.74$ & $32.80$ & $39.27$ \\
    GPS-Net$^{\dagger}$[19]   & $35.26$ & $74.81$ & $32.85$ & $33.98$ & $41.69$ \\
    GPS-Net$^{\dagger *}$[19]   & $38.93$ & $74.74$ & $32.77$ & $33.87$ & $41.60$ \\
    BGNN[16]  & $40.45$ & $74.98$ & $33.51$ & $34.15$ & $42.06$ \\
    \textbf{DR-IWSL} & $\mathbf{41.00}$ & $\mathbf{75.21}$ & $\mathbf{34.20}$ & $\mathbf{35.28}$ & $\mathbf{42.76}$ \\
	\bottomrule
    \end{tabular}
    \begin{tablenotes}
	\item [\textbullet] Note: All the above methods apply ResNeXt-101-FPN as the backbone. $*$ means the re-sampling strategy [30] is applied in this method, and $\dagger$ depicts the results reproduced with the latest code from the authors. Using bold to represent the proposed method.
      \end{tablenotes}
    \end{threeparttable}
\end{table} 

\section{Conclusion}

To avoid underestimating the underlying complex posterior, in this paper, we propose a novel doubly reparameterized importance weighted structure learning (DR-IWSL) method, which replaces the classical ELBO with a tighter importance weighted lower bound as the variational inference objective. Such lower bound is computed via multiple samples drawn from a reparameterizable Gumbel-Softmax sampler. More importantly, we use a doubly reparameterized gradient estimator to reduce the variance of the associated derivatives, and employ a generic entropic mirror descent method, rather than the traditional message passing technique, to solve the resulting constrained variational inference task. The proposed DR-IWSL method is validated on two popular SGG benchmarks: Visual Genome and Open Images V6. It achieves the state-of-the-art detection performance on both benchmarks. Currently, we only employ a mean field variational Bayes framework to solve the SGG task, which implies no structural dependencies are considered within the variational distribution. Relaxing this issue would be our next target.

\begin{ack}
This work was supported in part by the U.K. Defence Science and Technology Laboratory, and in part by the Engineering and Physical Research Council (collaboration between U.S. DOD, U.K. MOD, and U.K. EPSRC through the Multidisciplinary University Research Initiative) under Grant EP/R018456/1.
\end{ack}

\section*{References}
\medskip

{
\small

[1] You, Q., Jin, H., Wang, Z., Fang, C.,\ Luo, J. \ (2016). Image captioning with semantic attention. {\it In Proceedings of the IEEE conference on computer vision and pattern recognition}, pp. 4651-4659.

[2] Rennie, S. J., Marcheret, E., Mroueh, Y., Ross, J.,\  Goel, V. \ (2017). Self-critical sequence training for image captioning. {\it In Proceedings of the IEEE conference on computer vision and pattern recognition}, pp. 7008-7024.

[3] Yang, X., Tang, K., Zhang, H.,\ Cai, J. \ (2019). Auto-encoding scene graphs for image captioning. {\it In Proceedings of the IEEE/CVF Conference on Computer Vision and Pattern Recognition}, pp. 10685-10694.

[4] Teney, D., Liu, L.,\  van Den Hengel, A. \ (2017). Graph-structured representations for visual question answering. {\it In Proceedings of the IEEE conference on computer vision and pattern recognition}, pp. 1-9.

[5] Anderson, P., He, X., Buehler, C., Teney, D., Johnson, M., Gould, S.,\  Zhang, L. \ (2018). Bottom-up and top-down attention for image captioning and visual question answering. {\it In Proceedings of the IEEE conference on computer vision and pattern recognition}, pp. 6077-6086.

[6] Shi, J., Zhang, H.,\  Li, J. \ (2019). Explainable and explicit visual reasoning over scene graphs. {\it In Proceedings of the IEEE/CVF Conference on Computer Vision and Pattern Recognition}, pp. 8376-8384.

[7] Wainwright, M. J., \ Jordan, M. I. \ (2008). Graphical models, exponential families, and variational inference. {\it Foundations and Trends® in Machine Learning}, 1(1–2), 1-305.

[8] Fox, C. W., \ Roberts, S. J. \ (2012). A tutorial on variational Bayesian inference. {\it Artificial intelligence review}, 38(2), 85-95.

[9] Xu, D., Zhu, Y., Choy, C. B., \ Fei-Fei, L. \ (2017). Scene graph generation by iterative message passing. {\it In Proceedings of the IEEE conference on computer vision and pattern recognition}, pp. 5410-5419.

[10] Li, Y., Ouyang, W., Zhou, B., Wang, K., \ Wang, X. \ (2017). Scene graph generation from objects, phrases and region captions. {\it In Proceedings of the IEEE international conference on computer vision}, pp. 1261-1270.

[11] Dai, B., Zhang, Y., \ Lin, D. \ (2017). Detecting visual relationships with deep relational networks. {\it In Proceedings of the IEEE conference on computer vision and Pattern recognition}, pp. 3076-3086.

[12] Woo, S., Kim, D., Cho, D., \ Kweon, I. S. \ (2018). Linknet: Relational embedding for scene graph. {\it Advances in Neural Information Processing Systems}, 31.

[13] Yang, J., Lu, J., Lee, S., Batra, D., \ Parikh, D. \ (2018). Graph r-cnn for scene graph generation. {\it In Proceedings of the European conference on computer vision (ECCV)}, pp. 670-685.

[14] Wang, W., Wang, R., Shan, S., \ Chen, X. \ (2019). Exploring context and visual pattern of relationship for scene graph generation. {\it In Proceedings of the IEEE/CVF Conference on Computer Vision and Pattern Recognition}, pp. 8188-8197.

[15] Tang, K., Zhang, H., Wu, B., Luo, W., \ Liu, W. \ (2019). Learning to compose dynamic tree structures for visual contexts. {\it In Proceedings of the IEEE/CVF conference on computer vision and pattern recognition}, pp. 6619-6628.

[16] Li, R., Zhang, S., Wan, B., \ He, X. \ (2021). Bipartite graph network with adaptive message passing for unbiased scene graph generation. {\it In Proceedings of the IEEE/CVF Conference on Computer Vision and Pattern Recognition}, pp. 11109-11119.

[17] Li, Y., Ouyang, W., Zhou, B., Shi, J., Zhang, C., \ Wang, X. \ (2018). Factorizable net: an efficient subgraph-based framework for scene graph generation. {\it In Proceedings of the European Conference on Computer Vision (ECCV)}, pp. 335-351.

[18] Chen, T., Yu, W., Chen, R., \ Lin, L. \ (2019). Knowledge-embedded routing network for scene graph generation. {\it In Proceedings of the IEEE/CVF Conference on Computer Vision and Pattern Recognition}, pp. 6163-6171.

[19] Lin, X., Ding, C., Zeng, J., \ Tao, D. \ (2020). Gps-net: Graph property sensing network for scene graph generation. {\it In Proceedings of the IEEE/CVF Conference on Computer Vision and Pattern Recognition}, pp. 3746-3753.

[20] Liu, D., Bober, M., \ Kittler, J. \ (2021). Visual semantic information pursuit: A survey. {\it IEEE transactions on pattern analysis and machine intelligence}, 43(4), 1404-1422.

[21] Burda, Y., Grosse, R. B., \ Salakhutdinov, R. \ (2016). Importance Weighted Autoencoders. {\it In 4th International Conference on Learning Representations (ICLR)}.

[22] Jang, E., Gu, S., \ Poole, B. \ (2017). Categorical reparameterization with gumbel-softmax. {\it In 5th International Conference on Learning Representations (ICLR)}.

[23] Tucker, G., Lawson, D., Gu, S., \ Maddison, C. J. \ (2018). Doubly Reparameterized Gradient Estimators for Monte Carlo Objectives. {\it In 6th International Conference on Learning Representations (ICLR)}.

[24] Beck, A., \ Teboulle, M. \ (2003). Mirror descent and nonlinear projected subgradient methods for convex optimization. {\it Operations Research Letters}, 31(3), 167-175.

[25] Qi, M., Li, W., Yang, Z., Wang, Y., \ Luo, J. \ (2019). Attentive relational networks for mapping images to scene graphs. {\it In Proceedings of the IEEE/CVF Conference on Computer Vision and Pattern Recognition}, pp. 3957-3966.

[26] Zellers, R., Yatskar, M., Thomson, S., \ Choi, Y. \ (2018). Neural motifs: Scene graph parsing with global context. {\it In Proceedings of the IEEE conference on computer vision and pattern recognition}, pp. 5831-5840.

[27] Chawla, N. V., Bowyer, K. W., Hall, L. O., \ Kegelmeyer, W. P. \ (2002). SMOTE: synthetic minority over-sampling technique. {\it Journal of artificial intelligence research}, 16, 321-357.

[28] Shen, L., Lin, Z., \ Huang, Q. \ (2016). Relay backpropagation for effective learning of deep convolutional neural networks. {\it In European conference on computer vision (ECCV)}, pp. 467-482.

[29] Mahajan, D., Girshick, R., Ramanathan, V., He, K., Paluri, M., Li, Y., ... \ Van Der Maaten, L. \ (2018). Exploring the limits of weakly supervised pretraining. {\it In Proceedings of the European conference on computer vision (ECCV)}, pp. 181-196.

[30] Gupta, A., Dollar, P., \ Girshick, R. \ (2019). LVIS: A dataset for large vocabulary instance segmentation. {\it In Proceedings of the IEEE/CVF conference on computer vision and pattern recognition}, pp. 5356-5364.

[31] Hu, X., Jiang, Y., Tang, K., Chen, J., Miao, C., \ Zhang, H. \ (2020). Learning to segment the tail. {\it In Proceedings of the IEEE/CVF Conference on Computer Vision and Pattern Recognition}, pp. 14045-14054.

[32] Gidaris, S., \ Komodakis, N. \ (2018). Dynamic few-shot visual learning without forgetting. {\it In Proceedings of the IEEE conference on computer vision and pattern recognition}, pp. 4367-4375.

[33] Zhou, B., Cui, Q., Wei, X. S., \ Chen, Z. M. \ (2020). Bbn: Bilateral-branch network with cumulative learning for long-tailed visual recognition. {\it In Proceedings of the IEEE/CVF conference on computer vision and pattern recognition}, pp. 9719-9728.

[34] Guo, Y., Gao, L., Wang, X., Hu, Y., Xu, X., Lu, X., ... \ Song, J. \ (2021). From general to specific: Informative scene graph generation via balance adjustment. {\it In Proceedings of the IEEE/CVF International Conference on Computer Vision}, pp. 16383-16392.

[35] Cao, K., Wei, C., Gaidon, A., Arechiga, N., \ Ma, T. \ (2019). Learning imbalanced datasets with label-distribution-aware margin loss. {\it Advances in neural information processing systems}, 32.

[36] Cui, Y., Jia, M., Lin, T. Y., Song, Y., \ Belongie, S. \ (2019). Class-balanced loss based on effective number of samples. {\it In Proceedings of the IEEE/CVF conference on computer vision and pattern recognition}, pp. 9268-9277.

[37] Tang, K., Niu, Y., Huang, J., Shi, J., \ Zhang, H. \ (2020). Unbiased scene graph generation from biased training. {\it In Proceedings of the IEEE/CVF conference on computer vision and pattern recognition}, pp. 3716-3725.

[38] Zhang, J., Shih, K. J., Elgammal, A., Tao, A., \ Catanzaro, B. \ (2019). Graphical contrastive losses for scene graph parsing. {\it In Proceedings of the IEEE/CVF Conference on Computer Vision and Pattern Recognition}, pp. 11535-11543.

[39] Ren, S., He, K., Girshick, R., \ Sun, J. \ (2015). Faster r-cnn: Towards real-time object detection with region proposal networks. {\it Advances in neural information processing systems}, 28.

[40] Sutton, C., \ McCallum, A. \ (2012). An introduction to conditional random fields. {\it Foundations and Trends® in Machine Learning}, 4(4), 267-373.

[41] Scarselli, F., Gori, M., Tsoi, A. C., Hagenbuchner, M., \ Monfardini, G. \ (2008). The graph neural network model. {\it IEEE transactions on neural networks}, 20(1), 61-80.

[42] Gilmer, J., Schoenholz, S. S., Riley, P. F., Vinyals, O., \ Dahl, G. E. \ (2017). Neural message passing for quantum chemistry. {\it In International conference on machine learning}, pp. 1263-1272.

[43] Wang, X., Girshick, R., Gupta, A., \ He, K. \ (2018). Non-local neural networks. {\it In Proceedings of the IEEE conference on computer vision and pattern recognition}, pp. 7794-7803.

[44] Zhou, J., Cui, G., Hu, S., Zhang, Z., Yang, C., Liu, Z., ... \ Sun, M. \ (2020). Graph neural networks: A review of methods and applications. {\it AI Open}, 1, 57-81.

[45] Teboulle, M. \ (1992). Entropic proximal mappings with applications to nonlinear programming. {\it Mathematics of Operations Research}, 17(3), 670-690.

[46] Raskutti, G., \ Mukherjee, S. \ (2015). The information geometry of mirror descent. {\it IEEE Transactions on Information Theory}, 61(3), 1451-1457.

[47] Eicke, B. \ (1992). Iteration methods for convexly constrained ill-posed problems in Hilbert space. {\it Numerical Functional Analysis and Optimization}, 13(5-6), 413-429.

[48] Krishna, R., Zhu, Y., Groth, O., Johnson, J., Hata, K., Kravitz, J., ... \  Fei-Fei, L. \ (2017). Visual genome: Connecting language and vision using crowdsourced dense image annotations. {\it International journal of computer vision}, 123(1), 32-73.

[49] Kuznetsova, A., Rom, H., Alldrin, N., Uijlings, J., Krasin, I., Pont-Tuset, J., ... \ Ferrari, V. \ (2020). The open images dataset v4. {\it International Journal of Computer Vision}, 128(7), 1956-1981.

[50] Liu, Z., Miao, Z., Zhan, X., Wang, J., Gong, B., \ Yu, S. X. \ (2019). Large-scale long-tailed recognition in an open world. {\it In Proceedings of the IEEE/CVF Conference on Computer Vision and Pattern Recognition}, pp. 2537-2546.

[51] He, K., Zhang, X., Ren, S., \ Sun, J. \ (2016). Deep residual learning for image recognition. {\it In Proceedings of the IEEE conference on computer vision and pattern recognition}, pp. 770-778.

}

\section*{Checklist}

The checklist follows the references.  Please
read the checklist guidelines carefully for information on how to answer these
questions.  For each question, change the default \answerTODO{} to \answerYes{},
\answerNo{}, or \answerNA{}.  You are strongly encouraged to include a {\bf
justification to your answer}, either by referencing the appropriate section of
your paper or providing a brief inline description.  For example:
\begin{itemize}
  \item Did you include the license to the code and datasets? \answerYes{See Section~\ref{gen_inst}.}
  \item Did you include the license to the code and datasets? \answerNo{The code and the data are proprietary.}
  \item Did you include the license to the code and datasets? \answerNA{}
\end{itemize}
Please do not modify the questions and only use the provided macros for your
answers.  Note that the Checklist section does not count towards the page
limit.  In your paper, please delete this instructions block and only keep the
Checklist section heading above along with the questions/answers below.

\begin{enumerate}

\item For all authors...
\begin{enumerate}
  \item Do the main claims made in the abstract and introduction accurately reflect the paper's contributions and scope?
    \answerYes{}
  \item Did you describe the limitations of your work?
    \answerYes{}
  \item Did you discuss any potential negative societal impacts of your work?
    \answerNA{}
  \item Have you read the ethics review guidelines and ensured that your paper conforms to them?
    \answerYes{}
\end{enumerate}

\item If you are including theoretical results...
\begin{enumerate}
  \item Did you state the full set of assumptions of all theoretical results?
    \answerYes{}
        \item Did you include complete proofs of all theoretical results?
    \answerYes{}
\end{enumerate}

\item If you ran experiments...
\begin{enumerate}
  \item Did you include the code, data, and instructions needed to reproduce the main experimental results (either in the supplemental material or as a URL)?
    \answerNo{}
  \item Did you specify all the training details (e.g., data splits, hyperparameters, how they were chosen)?
    \answerYes{}
        \item Did you report error bars (e.g., with respect to the random seed after running experiments multiple times)?
    \answerNA{}
        \item Did you include the total amount of compute and the type of resources used (e.g., type of GPUs, internal cluster, or cloud provider)?
    \answerYes{}
\end{enumerate}

\item If you are using existing assets (e.g., code, data, models) or curating/releasing new assets...
\begin{enumerate}
  \item If your work uses existing assets, did you cite the creators?
    \answerYes{}
  \item Did you mention the license of the assets?
    \answerNA{}
  \item Did you include any new assets either in the supplemental material or as a URL?
    \answerNA{}
  \item Did you discuss whether and how consent was obtained from people whose data you're using/curating?
    \answerNA{}
  \item Did you discuss whether the data you are using/curating contains personally identifiable information or offensive content?
    \answerNA{}
\end{enumerate}

\item If you used crowdsourcing or conducted research with human subjects...
\begin{enumerate}
  \item Did you include the full text of instructions given to participants and screenshots, if applicable?
    \answerNA{}
  \item Did you describe any potential participant risks, with links to Institutional Review Board (IRB) approvals, if applicable?
    \answerNA{}
  \item Did you include the estimated hourly wage paid to participants and the total amount spent on participant compensation?
    \answerNA{}
\end{enumerate}

\end{enumerate}





\end{document}